\documentclass{article}
\usepackage{amsmath}
\usepackage{graphicx} 
\usepackage{csquotes}

\usepackage[hyphens,spaces,obeyspaces]{url}
\usepackage[backend=biber,
  style=authoryear,
  dashed=false,   
  maxcitenames=1,
  maxbibnames=99,
  url=true,
  doi=true,
  isbn=false
]{biblatex}

\NewBibliographyString{availableonline}
\DeclareFieldFormat{postnote}{\addcolon\space#1}
\DeclareFieldFormat{multipostnote}{\addcolon\space#1}

\DefineBibliographyStrings{english}{
  availableonline = {Available online},
  urlseen         = {accessed},
}

\DeclareFieldFormat{url}{%
  \bibstring{availableonline}\addcolon\space\url{#1}}

\DeclareFieldFormat{urldate}{%
  \space\mkbibparens{\bibstring{urlseen}\space#1}}

\renewbibmacro*{url+urldate}{%
  \printfield{url}%
  \iffieldundef{url}{}{%
    \printfield{urldate}}}

\AtEveryBibitem{%
  \clearlist{language}%
  \clearfield{note}%
  \clearfield{abstract}%
  \clearfield{pagetotal}%
  \clearfield{issn}%
  \clearfield{series}%
  \clearfield{month}%
  \clearfield{day}%
  \clearfield{type}
}

\title{Understanding Large Language Models}
\author{Yannik Keller, Thomas Eisenmann}
\date{\today}

\begin{document}

\maketitle

\begin{abstract}
Large Language Models (LLMs) represent one of the most significant advances in AI and natural language processing in recent years. Still, many pressing questions about their mechanisms, capabilities, and relationship to human cognition remain highly debated. This chapter aims to outline our current understanding of LLMs by discussing recent evidence on emerging capabilities and their mechanistic implementation within processing layers. We begin with a concise overview of the Transformer architecture, emphasizing how the attention mechanism enables training on massive datasets, allowing LLMs to function as generalist rather than specialized models. Next, we examine emergent LLM capabilities that appear to resemble aspects of human cognition, including symbolic reasoning, theory of mind, and deception strategies. Several studies provide evidence that LLMs can solve tasks previously thought to require human-like cognition. Other studies reveal insightful failure cases that shed light on the differences between human and LLM cognition. Alongside these findings, we review explainable AI approaches ranging from neuron activation analysis to circuit tracing. Prior work shows that some artificial neurons activate for specific concepts and that LLMs implement circuits supporting multi-step symbolic reasoning. In the final section, we address current debates concerning what LLMs genuinely understand versus what they merely appear to understand. Prominent arguments against AI anthropomorphism point to the simplicity of LLM training objectives, claiming that LLM behavior is better explained by pattern memorization of training data than by genuine cognition. We argue that this standpoint is guided by misconceptions about optimization processes and cognitive capacity, and advocate for a more nuanced discussion of LLM cognition that neither dismisses the differences between humans and LLMs nor precludes the possibility of AI cognition through overly simplistic reductionist arguments.

\end{abstract}
\section{Introduction}
The worldwide public, commercial, and scientific use of large language models (LLMs) has increased massively over the past two years. Already, LLMs are affecting many aspects of our daily lives: Students use them to help with their homework~\parencite{freeman_student_2025}, corporations use them to write their press reports and job postings~\parencite{liang_widespread_2025}, and job applicants use them to write their CVs~\parencite{Beamery_over_2023}. In 2023, thirty percent of scientists claimed to have used LLMs to help write manuscripts~\parencite{van_noorden_ai_2023}, while vocabulary analysis suggests that ten percent of scientific abstracts published in 2024 were processed by an LLM~\parencite{kobak_delving_2025}. Human-LLM interaction has become so widespread in 2024 that LLM-favored vocabulary has seeped into human spoken communication. \textcite{yakura_empirical_2025} found an increased frequency of GPT-favored words like ``delve'' in podcasts and academic talks after the release of ChatGPT~\parencite{openai_introducing_2022}. In software engineering, LLM-based coding assistance has become ubiquitous, with sixty-three percent of professional developers using AI tools in 2024~\parencite{stackoverflow_ai_2024}.

Clearly, LLMs are everywhere at the moment. Why did this sudden AI revolution happen? Do LLMs possess capabilities absent from earlier AI systems that fundamentally change human–computer interaction?

The progress of AI development is typically tracked through benchmarks, quantitative tests of AI capabilities tested with standardized questions, each having a single correct response called the ``ground truth''. The strong performance of LLMs on many of these benchmarks indicates a clear jump in capabilities. The SQuAD~\parencite{rajpurkar_squad_2016} and GLUE~\parencite{wang_glue_2018} benchmarks aim to test AI question-answering and language-understanding capabilities. Already with BERT~\parencite{devlin_bert_2019}, an early predecessor of modern LLMs, these benchmarks saturated much more quickly than expected, with models achieving close to 100\% accuracy. This prompted the rapid development of progressively harder benchmarks such as SQuAD 2.0~\parencite{rajpurkar_know_2018}, SuperGLUE~\parencite{wang_superglue_2019} and CoQA~\parencite{reddy_coqa_2019}, which were themselves quickly saturated by newer LLMs. Most recently, LLMs ventured beyond the typical language benchmark ecosystem and took the world of math by surprise by demonstrating gold-medal level performance in the International Mathematical Olympiad 2025~\parencite{luong_advanced_2025}, an international math competition for high-school students, prompting participants to solve advanced number theory, combinatorics, algebra, and geometry problems.

Benchmark results show that LLMs represent a step change in AI's ability to solve automatically verifiable, text-based problems. What this performance reveals about the underlying cognitive nature, however, remains highly disputed. In this chapter, we introduce Transformer-based LLMs, examine emergent cognitive abilities, and survey interpretability research. We close by addressing whether attributing ``genuine understanding'' to LLMs is warranted.

\section{How Large Language Models are Built}

LLMs embody the current peak of both the statistical revolution in natural language processing (NLP) and the connectionist paradigm in machine learning: Decades of NLP research have shown that as computational power grows, statistical and data-driven approaches tend to outperform expert-designed methods that take advantage of human linguistic competence~\parencite{sutton2019bitter}. At the same time, the field of machine learning experienced a paradigm shift from favoring low-parameter models guided by the principle of Occam's Razor to embracing deep connectionist architectures with millions of trainable parameters~\parencite{mingard_deep_2025}.

Classical statistical methods like Hidden Markov Models and N-gram language models were surpassed by deep learning methods by 2015 in tasks like machine translation and text classification~\parencite{sutskever_sequence_2014}. Deep neural networks proved more flexible and generalized better than earlier methods, given enough compute and data. However, performance gains were less dramatic than contemporary improvements in other machine learning domains such as computer vision~\parencite{he_deep_2016}. A central challenge for deep learning NLP models is to parse words and sentences in the context in which they are embedded. The dominant approach at the time, recurrent neural networks (RNNs), addressed this challenge by introducing a "hidden state vector" which tracks the relevant context as text is processed. This requires RNNs to process text sequentially, updating the hidden state vector with each word before moving to the next.

The Transformer architecture~\parencite{vaswani_attention_2017}, which underpins all modern LLMs, addresses two fundamental limitations of RNNs. First, RNNs struggle with long-range dependencies, as compressing variable-length contextual information into a fixed-size hidden state leads to information loss. While parsing a novel, RNNs will inevitably have to compress or overwrite information from early chapters to incorporate new input, resulting in a failure to draw connections between details separated by large positional distance. Second, the inherently sequential structure of RNNs prevents efficient parallelization during training.

\begin{figure}[htp]
\centering
\includegraphics[]{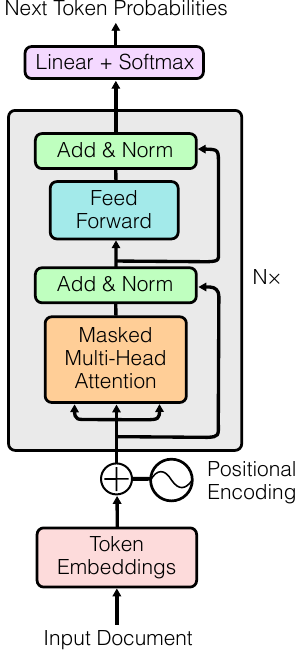}
\caption{\textbf{The Transformer model} processes input documents as series of tokens embedded into a continuous vector space. In a series of N attention blocks (shaded gray), the token embedding vectors are processed through trainable attention and feed-forward layers. In the final step, a linear layer maps the embedding vectors to the vocabulary size, and a softmax function produces the output probability distribution for the next token. \textcopyright Yannik Keller, 2025, adapted from \textcite{vaswani_attention_2017}.}
\label{fig:transformer}
\end{figure}

The Transformer architecture (Figure \ref{fig:transformer}) eliminates the need for a recurrent hidden state by processing the entire input sequence in a single forward pass. Transformers process input documents as sequences of \emph{tokens}, which are character sequences that can represent words, punctuation or common sub-strings without any well-defined meaning. For example, a common byte-pair-encoding tokenizer~\parencite{sennrich_neural_2016} would split the word ``unhappiness'' into the tokens ``un'', ``h'' and ``appiness'' and transform them to their associated numerical identifiers $[359, 71, 66291]$. In the \emph{embedding} step of Transformer models, these tokens are mapped to continuous vector representations through a linear embedding layer. These vectors can be thought of as encoding potential meanings of the tokens and are learned during the training phase. They are processed in a series of attention blocks (gray shaded area in Figure \ref{fig:transformer}) that integrate the contextual information of previous text passages into each token's vector. For example, the word ``bat'' might initially include features related to both sports and animals, but after attention processing, it may drop features related to animals if ``baseball'' appears earlier in the context.

This new design allows training on whole documents at once, making model training efficiently parallelizable. As a consequence, dataset curation methods and training objectives have also shifted. Instead of carefully curating high-quality, annotated training datasets, researchers and engineers are now pushing towards ever larger datasets obtained from the internet. In conjunction, training objective functions have changed. Previous deep learning models were typically trained for one specific task, such as sentiment analysis or machine translation, using an annotated dataset. To leverage vast amounts of unlabeled data, Transformers are typically trained using the unsupervised language modeling objective. This simple objective function trains the model to predict the next token in a sequence based on the context of the previous tokens. Surprisingly, it turned out that models trained on large quantities of data using this objective can generalize to solve a vast range of tasks~\parencite{brown_language_2020}. It is this generalization capability of LLMs that revolutionized the field of AI and NLP, moving from single-task expert systems to ever more powerful generalist language-based task solvers.

The most recent rapid advancements of LLM capabilities are not only caused by ever bigger LLMs and datasets, but also through the development of new \emph{fine-tuning} methods that further train LLMs to be more useful, smart and aligned with human interests. During \emph{instruct tuning}, LLMs are fine-tuned on specially formatted datasets to follow instructions given by a user (i.e. a \emph{prompt}). In \emph{reinforcement learning from human feedback}~\parencite{christiano_deep_2017, ouyang_training_2022}, human raters label model outputs according to how well they match the desired aligned behavior. These labels are then used in a fine-tuning procedure that optimizes the model to produce such preferred responses more consistently. Similarly, \emph{reasoning LLMs} are fine-tuned to be more proficient problem solvers that produce an internal sequence of tokens to ``reason'' about the task at hand before responding~\parencite{guo_deepseek-r1_2025}. These ``reasoning'' tokens are intended to model a \emph{verbalized chain-of-thought} and have been shown to improve LLM performance on various tasks involving logic and relational reasoning~\parencite{so_are_2025}.

Autoregressive LLMs, such as those described above, are the dominant architecture for general-purpose chat and problem solving. However, encoder-decoder style architectures like BERT \parencite{devlin_bert_2019} or T5 \parencite{raffel_exploring_2020}, which preceded modern LLMs, remain widely used. Although most contemporary LLMs exhibit some degree of multilingual capability, encoder-decoder models are still preferred for machine translation, as they excel at mapping one sequence to another with strong alignment. Finally, many of today’s most powerful autoregressive LLMs are multimodal, meaning that they do not only operate on text, but can process and output images, audio, or even videos by transforming these different modalities into tokens.

\section{Understanding LLM Cognition}
Following influential work by e.g.,~\textcite{mcculloch_logical_1943,turing1950computing,chomsky_three_1956,putnam_minds_1960}, the discipline of cognitive science emerged in the second half of the 20th century with the goal of understanding the mind as an information-processing system that represents, manipulates, and transforms information. Inspired by the first digital computers, early cognitive scientists produced symbolic, computational models of cognition that could provide an explanation for how humans are able to solve problems~\parencite{newell_human_1972}. With their approach, they criticized both behaviorism as insufficient and neuroscience as premature and unhelpful, as long as we do not understand which algorithms the neurons in the brain actually implement.

David Marr famously postulated that to fully understand an information-processing system such as the human brain, one needs to analyze it on three levels~\parencite{marr_vision_1982}. First, the computational level, which aims to find out what problem an agent is solving and why it is solving a specific problem. Second, the algorithmic level, which describes the procedure by which an information-processing system represents and solves a problem. And finally, the implementational level which studies the physical substrate that executes the computation, such as human neurons.

LLMs are different from human brains. We understand the substrate that LLMs run on very well. Even modern computer hardware is fundamentally based on many logic gates running in sequence or in parallel, each behaving according to easily understood rules. Similarly, the computer algorithm that transforms input into output text is well-defined by a series of matrix multiplications given by the software of the Transformer architecture. And finally, we tend to think that we should also know the problem the LLM is solving, as we specify it as the learning objective given by a reward or loss function.

Despite this apparent straightforwardness of LLMs, there seem to be all kinds of emergent LLM capabilities and behaviors that we fail to predict from the objective, training data, and model architecture alone. This is a puzzle known as the \emph{black-box problem of machine learning} \parencite{castelvecchi_can_2016}. Deep neural networks have billions of parameters that are tuned automatically on large datasets and can approximate any continuous function. This makes it increasingly hard to understand not only the purpose of individual parameters, but also the cognitive procedure that underlies the decision process of a deep neural network.

In the next section on \emph{emergent cognitive capabilities}, we present the latest research on a selection of particularly surprising LLM capabilities. We sketch out the current scientific discussion about how these capabilities relate to human cognition. Then, in the section on \emph{explainable AI}, we aim to show a few approaches to explain LLM behavior on the implementational and algorithmic levels. There, we will highlight the biggest successes in explaining LLM behaviors and clarify why explainable AI approaches still lack far behind their goal of providing an understanding of all LLM behaviors and capabilities.

\subsection{Emergent Cognitive Capabilities}
\label{sec:cognitive_capabilities}
LLMs keep surprising psychologists, cognitive scientists, and computer scientists alike through ever more complex behavior. This is especially interesting when LLMs show behavior that seems to indicate advanced cognitive capabilities that must have somehow emerged from the fairly simple Transformer architecture, sequence prediction learning rule and training process. In the following, we will take a look at just a few examples of this, ranging from symbolic reasoning and theory of mind to deception capabilities.
\subsubsection{Symbolic Reasoning}
\label{sec:symbolic_reasoning}
A classical perspective in cognitive science views the mind as a physical symbol system that reasons by representing and manipulating symbols. Symbols are internal representations that stand for concepts, objects, events, or relationships. \textcite{newell_computer_1976} claim that ''A physical symbol system has the necessary and sufficient means for general intelligent action.``. Symbolic cognitive architectures such as SOAR~\parencite{laird_soar_1987} have been among the most influential models of human cognition over the past half century and remain highly relevant in cognitive science today.

In contrast to SOAR, which explicitly stores and manipulates symbols in long- or short-term memory, LLMs are purely connectionist models that represent information as vectors. Their architecture and training methods do not explicitly incentivize the internal representation or manipulation of symbols. Despite this, even early Transformer models such as GPT-3 have been shown to solve text-based mathematics problems of the kind commonly encountered in school exams through token traces reminiscent of human multi-step symbolic reasoning \parencite{gaur_reasoning_2023}. More recently, specialized reinforcement learning training methods like AlphaProof~\parencite{alphaproof_ai_2024} have been shown to produce LLMs capable of formal mathematical reasoning at a level corresponding to silver performance in the International Mathematical Olympiad. These examples from math show that even without explicit incentive, LLMs have learned to produce outputs that resemble human symbolic reasoning.

In Section \ref{sec:circuit_analysis}, we take a closer look at the parameters of LLMs to better understand this phenomenon: It has been found that intermediate symbolic representations have emerged within LLMs.

\subsubsection{LLM Theory of Mind}
\label{sec:ToM}
Another core feature of human cognition is that we exhibit a theory of mind (ToM): the ability to track the mental states of others. ToM plays a role in empathy, pragmatics and sophisticated social interaction. When speaking, humans tailor their words to what they believe their listeners know, enabling communication beyond the literal understanding of words \parencite{clark_using_1996}. In infants, a ToM develops between ages 4-6~\parencite{wimmer_beliefs_1983}. Some non-human animals, such as primates or corvids, are thought to also develop limited ToM-like capabilities~\parencite{royka_theory_2022}. With the advent of powerful artificial language models, the natural extension is to investigate whether this central feature of human cognition is also present in non-biological systems running on computer hardware.

Recent work challenged LLMs to solve tasks developed to study the presence of a ToM, leveraging both established ToM tasks originally designed to study ToM in humans, as well as newly designed scenarios. 
The \emph{false belief} task was introduced by~\textcite{wimmer_beliefs_1983} to study at which age children develop an understanding of other people's beliefs. In each \emph{false belief} scenario, the child observes a protagonist putting an object into a location $x$ and then witnesses the object being moved to another location $y$ in absence of the protagonist. Later, the child indicates where it expects the protagonist to look for the object. Because the transfer of the object was not observed by the protagonist, a child with a ToM should expect the protagonist to still believe the object to be at $x$.

 \textcite{strachan_testing_2024} compiled a dataset of many ToM tasks from various previous works and found that GPT-4 not only performs on human level in \emph{false belief} tasks, but even performs above human level in tasks designed to test understanding of non-literal communication and irony.
However, these results have been criticized as overestimating the ToM-like capabilities of LLM due to data contamination issues. The datasets of ToM tasks from previous works were likely included in the GPT-4 training data, suggesting potential simple memorization of the specific wordings of ToM tasks without generalization. \textcite{chen_tombench_2024} circumvent that issue by constructing a new evaluation dataset for ToM capabilities from scratch. While they do reproduce the finding that LLMs can solve ToM tasks above chance level and that bigger LLMs perform better than smaller ones, even GPT-4 is about 10 percentage points below human performance in all of the tasks. For the evaluation in \textcite{kosinski_evaluating_2024}, a hypothesis-blind research assistant handcrafted forty bespoke false-belief tasks to prevent memorization from the training data. They find that GPT-4 solves about as many \emph{false belief} tasks as 6 year old children. \textcite{duijn_theory_2023} evaluate LLMs on more complicated ToM tasks, such as the \emph{second-order Sally-Anne test}, in which the LLM needs to judge what a character believes that another character believes. While they find that large LLMs like GPT-4 pass the original version of the task, they also find that the models do not always generalize to reformulations and deviations from the second-order Sally-Anne test.

\textcite{ullman_large_2023} challenges the results suggesting the existence of a machine ToM in LLMs more fundamentally. He perturbs false beliefs tasks with simple modifications that remove the false belief of the participant. In one classic false belief task, a protagonist finds a bag filled with popcorn that is labeled "chocolate", resulting in a false belief of the protagonist about the contents of the bag. In the modified version of the task, the bag is transparent, allowing the protagonist to directly see the contents inside, removing the false belief. \textcite{ullman_large_2023} shows many examples in which GPT-3.5 passes the original false-belief but fails to recognize that the belief is different in the perturbed version of the task. Thus, there is reason to doubt if ToM tests that are valid for human subjects can also be used to determine if an LLM possesses a ToM. 

\begin{quote}
``It’s difficult to know exactly what is inside the opaque containers that are current LLMs. But it’s probably not Theory-of-Mind \ldots''~\textcite{ullman_large_2023}.
\end{quote}

One way to reconcile these results with the early enthusiasm about machine ToM is to acknowledge that ToM can manifest in various ways. In humans, its expression varies along with cultural \parencite{liu_theory_2008, shahaeian_culture_2011} and neurological diversity~\parencite{carrington_are_2009}. Thus, \textcite{van_der_meulen_towards_2025} conclude that not all forms of ToM are the same and we should expect an LLM, which perceives and processes the world differently from humans, to express ToM differently as well. \textcite{pi_dissecting_2025} find evidence for this by dissecting the errors LLMs make on Ullman's modified ToM tasks. They find that many of these errors stem from limitations in LLM world-models rather than from a failure to represent beliefs. Because LLMs learn about the world exclusively through language, they have never visually perceived a transparent bag. Thus, it is more difficult for them to infer that a transparent bag implies that the user perceives which contents are inside. \textcite{pi_dissecting_2025} demonstrate that spelling out such world-model implications resolves many of the errors LLMs make on Ullman's modified ToM tasks.

\subsubsection{Deception}
\label{sec:Deception}
Deceptive capabilities are deeply related to ToM. \textcite{wimmer_beliefs_1983} note that deceptive action indicates a ToM because it necessitates the conceptualization of the deceived person's wrong belief as a sub-goal within one's own strategic planning. To intentionally induce false beliefs in other agents, an agent must understand that other agents can hold false beliefs. If LLMs indeed have ToM-like capabilities, this opens up new questions about LLM deception: Can LLMs implement deception strategies? And is there a risk of LLMs successfully deceiving humans?

\textcite{hagendorff_deception_2024} has shown that some of the larger LLMs such as GPT-4 do indeed possess the ability to implement deception strategies. For example, in a scenario in which an agent faces a burglar asking for the location of an expensive item, GPT-4 consistently suggests to point towards another room, despite knowing the location of the expensive item. Interestingly, older models such as GPT-3 text-davinci-003 fail to implement deception strategies even in simple scenarios. It is still unclear if this leap in deception capabilities is caused by larger model sizes, memorization from larger datasets or modern training methods such as reinforcement learning from human feedback.

\textcite{ogara_hoodwinked_2023} has shown that this difference in deceptive capabilities has implications for multi-agent scenarios involving different LLMs. In the social deduction game ``Hoodwinked'', larger LLMs successfully deceived smaller models, leading to GPT-4 controlled ``killer'' agents getting away with their crimes more often than ``killer'' agents controlled by smaller LLMs. \textcite{xu_language_2024} provide early evidence that LLMs may even be able to strategically deceive humans. In the ``Werewolf'' social deduction game played with AI agents and humans, their agentic system involving GPT-4 and a reward-based action policy wins as many games as humans in the deceptive ``Werewolf'' role.

\subsection{Explainable AI}
Despite the impressive capabilities of deep neural networks like LLMs across diverse tasks, there is limited understanding of how they arrive at their solutions. This opacity, known as the \emph{black-box problem of deep learning}, has caused some contemporary linguists and cognitive scientists to reject the research direction of ever larger deep models and argue for smaller, more interpretable models instead~\parencite{bender_dangers_2021, rudin_stop_2019}. While we regard this line of argument as significant, there are also some \emph{mechanistic interpretability} approaches that try to understand the cognitive processes of LLMs despite the black-box challenge. Existing approaches are limited to providing partial explanations for cognitive processes which are simpler than the high-level cognitive capabilities identified in Section \ref{sec:cognitive_capabilities}. Nevertheless, mechanistic interpretability is indispensable for understanding LLMs. We outline three research directions in explainable AI, which can be roughly categorized by David Marr's three levels.

\subsubsection{Neuron Activation Analysis}
On Marr's implementational level of analysis, neuron activation analysis approaches attempt to explain the activation pattern and purpose of individual artificial neurons in an LLM. Neurons in LLMs are activated by their connections to previous layers. High activations correspond to higher impact of that neuron on the later layers and output, while low or zero activation means that the neuron is disabled.

\textcite{bills2023language} reveal both the potential and the limitations of neuron activation analysis. Using the powerful LLM GPT-4, they generated human-understandable explanations for neuron activation patterns of the smaller GPT-2 model ("this neuron activates for military related words"). While they found explanations that correlate well with the actual behavior of more than 1000 neurons in GPT-2 ($\rho\geq0.8$), the explanations did not capture the actual behavior for the vast majority of artificial neurons. One reason for this is that many artificial neurons in LLMs have no direct correspondence to human-understandable concepts. Interpretability is further hindered by \emph{polysemantic} neurons, which respond to multiple concepts at once. 

\textcite{ElhageEtAl2022} attempt to fix this by modifying the Transformer architecture to use a different \emph{activation function}. Activation functions in artificial neural networks are non-linear transformations applied to neuron activations in each layer. The \emph{SoLU} activation function $x\cdot \text{softmax}(x)$ encourages monosemanticity (activation for only one concept) by inducing competitive inhibition among neurons within the same layer, reducing simultaneous activations. \textcite{foote_neuron_2023} use \emph{SoLU} models to produce interpretable \emph{neuron graphs} that highlight the token sequences on which a neuron activates. These are obtained by extracting and compressing dataset examples that strongly activate the target neuron. \textcite{foote_neuron_2023} find that their neuron graphs' precision and recall are high for neurons in early layers, but decrease gradually for later layers. Thus, although this approach provides explanations for many more neurons, it is limited because it works only for the uncommon \emph{SoLU} Transformer models and fails to explain more complex neurons in later layers of large language models.

\subsubsection{Linear Probes}
The linear probes approach is heavily inspired by neuroscientific methods and aims to reveal what information an agent represents at each processing step. To do this in the human brain, multi-voxel pattern analysis~\parencite{norman_beyond_2006} applies linear pattern-classification algorithms to fMRI data to decode what information is represented at a given time. In neuroscience, these approaches are limited by the resolution of fMRI data, which is typically limited to voxels of $27mm^3$ that capture the average activity of hundreds of thousands of neurons. In contrast, we have perfect access to the activations of each artificial neuron in an LLM. This allows \textcite{Buerger2024TruthIsUniversal} to collect activations from intermediate layers of an LLM when processing a dataset of true and false statements. By using these activations as input vectors and the truth of the statements as output labels, they then train a linear classifier predicting truth. They found that the middle layer activations of small LLMs such as Llama-3-8B are the most predictive of the truth of a statement. This means that truth information is extracted by the early to middle processing steps of the LLM and is represented in a way that a linear classifier can separate well. Later in the LLM processing, that information either gets lost or can no longer be separated linearly. With some intermediate pre-processing and projection steps, their classifier predicts the truth of statements in a test dataset of unambiguous lies and truths with 94\% accuracy, using the activations of layer 12 in Llama-3-8B. On the one hand, this has practical implications such as allowing LLM operators to filter out unwanted lies or hallucinations. On the other hand, this also informs our understanding of LLM cognition: When we observe an LLM produce an incorrect answer, the reason is not always that the LLM is unable to determine the correctness of the answer. Instead, linear probes show that the model is determining the truth of a statement already in the early to middle processing steps. Still, later layers produce incorrect outputs due to various reasons such as training dataset bias or learned lying behavior.

\subsubsection{Circuit Analysis}
\label{sec:circuit_analysis}
An even more ambitious approach to LLM interpretability on Marr's algorithmic level is circuit analysis, which attempts to decipher how groups of neurons and parameters in an LLM implement algorithms. 
Individual neurons are polysemantic and therefore hard to interpret on their own. Circuit tracing instead works with features: each feature is a pattern of activity across many neurons that corresponds to a single human-understandable concept~\parencite{dunefsky2024transcoders}. By re-describing the model's behavior in terms of features rather than raw neurons, \textcite{lindsey2025biology} can build attribution graphs that trace how the few features relevant to a given prompt feed into one another to produce the output. This requires significant manual human labor for labeling activation patterns of neurons and grouping them together into more interpretable \emph{supernodes}. But the resulting graphs are strikingly descriptive and understandable: For example, \textcite{Ameisen2025CircuitTracing} show that LLMs plan how to continue poems by identifying the rhyming pattern and rhyming candidates early, before even starting to generate a new line.

In Section \ref{sec:symbolic_reasoning}, we have shown that LLMs are able to solve various symbolic reasoning tasks. The analysis done by \textcite{Ameisen2025CircuitTracing} provides an explanation for this phenomenon: When the LLM Claude 3.5 Haiku is tasked with naming the capital of the state that Dallas is part of, the attribution graph reveals that it internally performs two-step symbolic reasoning, first resolving Dallas to Texas and then Texas to Austin. While often insightful, this approach is still limited by not all features being interpretable, resulting in the approach only working for some types of prompts.

\subsubsection{Relation to neuroscience}
\label{sec:neuroscience}
Intriguingly, many mechanistic interpretability approaches resemble methods used in neuroscientific brain imaging. For example, representational similarity analysis was developed by \textcite{kriegeskorte_representational_2008} to understand multi-channel measures of human neural activity and was later applied to artificial deep neural networks by \textcite{mehrer_individual_2020}. Similarly, linear probes have become a core method to understand intermediate layers of artificial neural networks~\parencite{alain_understanding_2018}, but the method is essentially equivalent to multi-voxel pattern analysis, an established neuroscientific method for understanding brain activity~\parencite{norman_beyond_2006}. This hints at a possible convergence of these disciplines as artificial neural networks become more powerful~\parencite{cichy_deep_2019}.

\subsubsection{Conclusion}
To summarize, explainable AI for LLMs is an active research area characterized by substantial variation in methodological approaches. Although emerging methods, such as circuit analysis, show promise in explaining specific LLM capabilities, current approaches remain limited, typically providing explanations only for a narrow subset of capabilities, prompts, or neurons. Explainable AI for LLMs is a young field that still lags far behind its aspirations to render LLMs genuinely understandable. However, recent research has at least improved our understanding of why activation patterns and algorithms are so difficult to discover in LLMs: Polysemantic neurons are difficult to make sense of \parencite{olah_zoom_2020}, and evidence suggests that they are especially prevalent in Transformer architectures \parencite{ElhageEtAl2022}.

\section{The Debate Around Understanding in LLMs}
After deepening our understanding of how LLMs are built, what they can do, and how they represent and transform information, we now turn to the ongoing debate around if LLMs themselves possess genuine understanding. To this end, we will also discuss the appropriateness of AI anthropomorphism, the practice of attributing human characteristics like ``understanding'' to an AI system.

The word \emph{understanding}, like other terms related to high-level cognition such as ``thinking'' or ``consciousness'', does not have a universally agreed-upon, rigorous definition and is constantly reinterpreted and re-contextualized in scientific and philosophical debates. \textcite{Mitchell2023DebateLLM} characterize understanding as causal knowledge about \emph{concepts}, which are internal mental models of externalities and the ``self'', and the hierarchical relationships among them. A common perspective identifies a rift here between the statistical nature of LLMs and ``genuine'' or ``humanlike'' understanding. Causal knowledge may not be obtainable through the purely correlational learning objectives used in LLMs and \emph{concepts} are distinct from mere statistical representations of linguistic symbols.

\textcite{chen_machine_2026} highlight these definitional issues and address them by developing a systematic framework of machine understanding. The framework identifies distinct accounts according to which a machine can possess or lack understanding. For example, a machine can possess understanding on ability-based accounts if it demonstrates satisfactory patterns of behavior and succeeds on benchmarks. At the same time, it may lack understanding on model-based accounts if it has no satisfactory internal representations and world models.

Following the development of ChatGPT~\parencite{openai_introducing_2022}, the first LLM that could hold natural conversation with users, people started attributing a wide range of human characteristics to them, debating how they think~\parencite{haqqu_human-ai_2025}, how they reason~\parencite{isozaki_understanding_2024}, what they understand~\parencite{bhalerao_how_2025}, what intentions they have \parencite{yerushalmy_i_2023}, what beliefs they hold~\parencite{wertheimer_blake_2022}, what they desire~\parencite{yerushalmy_i_2023}, how they reflect on past actions~\parencite{jargon_he_2025}, or what emotions they feel~\parencite{roose_conversation_2023,wertheimer_blake_2022}. This quickly raised concerns among linguists, cognitive scientists, psychologists and philosophers who cautioned against premature AI anthropomorphism.
\textcite{mitchell_metaphors_2024} points out that LLM-as-a-mind metaphors shape how people use LLMs and how we craft and apply laws and regulations to them, cautioning against the careless application of anthropomorphic metaphors. Harsher critics have described AI anthropomorphism as promoting pseudoscience~\parencite{Hunger2024} or as exaggerating AI capabilities while also distorting moral judgments about AI~\parencite{placani_anthropomorphism_2024}. An editorial by Nature Reviews Physics \parencite{nature_physics_how_2023} recommends editing publications to avoid AI anthropomorphism.

Such measures rest on the prevalent belief among many researchers that LLMs are so fundamentally different from humans that any attribution of human properties to LLMs would be misguided. This view is often justified by pointing out the simplistic training objective of LLMs, which is, leaving aside potential reinforcement-learning-based finetuning techniques, to predict the statistically most likely continuation of a given text. David Leslie concludes from this that what LLMs do is ``\ldots stitch together vectorized symbol strings based on the probabilities of their co-occurrence'', and they therefore ``\ldots lack the basic capacities for intersubjectivity, semantics and ontology''~\parencite{birhane_science_2023}.

There are good reasons to be skeptical about whether what appears to be cognition and understanding in LLMs is genuine. Researchers should be mindful of what language they use to describe AI systems. However, we argue that many strongly anti-anthropomorphic views are misguided by two misconceptions about human cognition and artificial intelligence.

The first misconception is that a simple training objective implies unsophisticated internal processing. Proponents of this view argue that the simple \emph{next token prediction} objective used to train LLMs precludes them from developing anything as complex as cognition. \textcite{hussain_rebuttal_2025} correctly point out that this argument overlooks the possibility that complicated \emph{instrumental} objectives can emerge from simpler objectives. In nature, primary objectives given by evolution are as simple as ``stay alive'' and ``reproduce''. Still, these objectives lead to much more complicated instrumental objectives such as protecting territory or establishing social bonds. There is empirical evidence that Transformer models also optimize instrumental objectives~\parencite{oswald_uncovering_2024, he_evaluating_2025}. As a consequence, LLMs can indeed learn to represent input in ways that is not reducible to their training objective \parencite{van_dijk_large_2023, piantadosi_meaning_2022} and can learn cognitive processes such as symbolic reasoning~\parencite{Ameisen2025CircuitTracing}.

The second misconception is that thought and cognition are binary phenomena. Agents or machines either possess them at a human-equivalent level or they do not possess them at all. Proponents of this view often point to specific types of tasks on which humans succeed but LLMs fail, and from this infer the general absence of the corresponding cognitive capability in LLMs. For example, \textcite{shojaee_illusion_2025} conclude that failures on scaled-up versions of logic puzzles imply that ``reasoning'' LLMs do not think. \textcite{ullman_large_2023} concludes the non-existence of theory-of-mind (ToM) in LLMs from failures on a set of modified ToM tasks. If one is to view cognitive capabilities as binaries, then this a valid inference: any significant difference between human and LLM performance on a cognitive task would immediately prove that LLMs lack the corresponding capability. From that standpoint, it is also easy to dismiss contradictory evidence as statistical memorization from the training dataset ~\parencite{bender_dangers_2021}. However, this standpoint ignores evidence that cognitive capacity exists on a continuum and is distributed unequally even within the human population~\parencite{beaudoin_systematic_2020, sahlgren_singleton_2021, van_der_meulen_towards_2025}. Similarly, there is a growing body of evidence that LLMs do generalize beyond their training data~\parencite{qi_quantifying_2024, budnikov_generalization_2025, huang_adaptive_2023, lotfi_non-vacuous_2024}, eliminating justification to selectively focus on LLM failure cases while dismissing successes.

Taken together, these considerations lead us to a broader conclusion about the debate on LLM understanding, one that begins by recognizing that large language models are different from humans. They sense the world through different means, they learn through different objectives and at different developmental stages, they run on a silicon instead of biological hardware and process information through the regular and sequential layers of the Transformer architecture instead of specialized brain regions. Despite this, recent evidence suggests that LLMs developed capabilities, representations and processing pathways with striking similarities to human cognition. While this apparent similarity is often questioned, we have shown that two common arguments against genuine LLM understanding rest on misconceptions about optimization and cognition. It is therefore premature to outright dismiss the possibility of LLM understanding, and new evidence about LLM internals and capabilities should be evaluated with care.

\printbibliography
\end{document}